\documentclass{article}

\usepackage[preprint]{neurips_2026}

\usepackage[utf8]{inputenc} 
\usepackage[T1]{fontenc}    
\usepackage{hyperref}       
\usepackage{url}            
\usepackage{booktabs}       
\usepackage{amsfonts}       
\usepackage{nicefrac}       
\usepackage{microtype}      
\usepackage{amsmath}
\usepackage{xcolor}         
\usepackage{comment}
\usepackage{graphicx}
\usepackage{pgfplots}
\usepackage{subcaption}
\pgfplotsset{compat=1.18}
\usepackage{multirow}
\usepackage{booktabs}
\usepackage{colortbl}

\title{KappaPlace: Learning Hyperspherical Uncertainty for Visual Place Recognition via Prototype-Anchored Supervision}

\author{%
  Maya Yanko \\
  Faculty of Engineering\\
  Bar-Ilan University\\
  Ramat Gan, Israel \\
   \And
  Yoli Shavit\thanks{Corresponding Author. Email: \texttt{yoli.shavit@biu.ac.il} } \\
  Faculty of Engineering\\
  Bar-Ilan University\\
  Ramat Gan, Israel \\
}

\begin{document}

\maketitle

\begin{abstract}
Visual Place Recognition (VPR) is critical for autonomous navigation, yet state-of-the-art methods lack well-calibrated uncertainty estimation. Standard pipelines cannot reliably signal when a query is ambiguous or a match is likely incorrect, posing risks in safety-critical robotics. We propose KappaPlace, a principled framework for learning uncertainty-aware VPR representations. Our core contribution is a Prototype-Anchored supervision strategy that leverages latent class representatives as targets for a probabilistic objective. By modeling image descriptors as von Mises-Fisher (vMF) variables, we learn a lightweight module to predict the concentration parameter $\kappa$ as a direct proxy for aleatoric uncertainty. While existing VPR uncertainty methods are typically restricted to a query-centric view, we derive a novel match-level formulation to quantify the reliability of specific query-reference pairs. Across five diverse benchmarks, KappaPlace reduces Expected Calibration Error (ECE@K) by up to 50\% compared to existing methods while maintaining or improving retrieval recall. We provide both a joint-training variant and a post-training extension for frozen backbones. Our results demonstrate that KappaPlace provides a robust, stable, and well-calibrated signal that enables reliable decision-making within the VPR pipeline. Our code is available at: \href{https://github.com/mayayank95/UncertaintyAwareVPR}{https://github.com/mayayank95/UncertaintyAwareVPR}. 
\end{abstract}

\section{Introduction}\label{sec:introduction}
Visual Place Recognition (VPR) provides the critical global localization required for autonomous navigation. Conventionally framed as a large-scale image retrieval task, VPR requires mapping a query image to a geographically tagged counterpart in a reference database. Driven by deep metric learning, contemporary pipelines achieve remarkable discriminative power by embedding images into high-dimensional latent manifolds~\cite{ali2023mixvpr, berton2022rethinking, berton2023eigenplaces}. However, while these models excel at identifying the "nearest neighbor" in feature space, they remain essentially deterministic. They provide point-estimates in embedding space but lack a native mechanism to express the certainty of those estimates. Beyond the lack of query-level uncertainty estimation, the match scores themselves (typically derived from cosine similarity or Euclidean distance) are uncalibrated; they represent relative proximity rather than likelihood of match validity.

Current attempts to bridge this gap suffer from several distinct limitations. Some frameworks are strictly coupled to specific training objectives, such as contrastive triplet losses \cite{warburg2021bayesian}, which restricts their utility in contemporary, scalable VPR paradigms like classification-based learning \cite{berton2022rethinking} or multi-similarity objectives \cite{Alibey_2022_gsvcities}. Crucially, these probabilistic adaptations often incur a measurable degradation in retrieval accuracy compared to their deterministic counterparts \cite{warburg2021bayesian}. Other approaches rely on restrictive score or pose distribution priors \cite{ miller2026through, zaffar2024estimation}. While effective for filtering, these methods output-score heuristics or characterize external environmental geometry rather than the aleatoric uncertainty inherent in the visual input itself. High-performing "post-hoc" solutions, such as geometric verification, offer reliability only at the cost of prohibitive computational burdens \cite{zaffar2024estimation}. Furthermore, techniques such as multi-pass Monte Carlo sampling primarily capture epistemic (model) uncertainty, failing to isolate data-dependent noise in a single forward pass and often yielding inferior calibration compared to specialized aleatoric models \cite{cai2022stun, warburg2021bayesian}. Finally, even recent self-teaching uncertainty networks (STUN) \cite{cai2022stun}, though theoretically architecture-agnostic, consistently degrade the performance of state-of-the-art VPR backbones, {as shown in Table \ref{tab:sf_final}}.

In this work, we present KappaPlace, a framework that provides well-calibrated uncertainty for both the query image and individual query-reference matches without compromising retrieval performance. The core of our approach is a Prototype-Anchored supervision strategy, which leverages stable latent cluster representatives as canonical targets for a probabilistic objective. We primarily demonstrate this strategy within classification-based VPR, where learned class embeddings serve as these anchoring targets. However, we show that this principle is fundamentally extensible to contrastive-based VPR by utilizing dynamic batch-level centroids. By anchoring uncertainty to these prototypes rather than noisy image-level pairs, KappaPlace achieves a more stable estimation of aleatoric noise. We model descriptors as von Mises-Fisher (vMF) variables on the unit hypersphere and optimize a lightweight auxiliary module to estimate the concentration parameter ($\kappa$) with a numerically stable bounding and approximation of the vMF loss~\cite{amos1974computation, sra2012short}. This enables the quantification of query-level uncertainty in a single forward pass, from which we further derive a novel match-level uncertainty score based on the $\kappa$-weighted resultant vector of the query and reference embeddings~\cite{jammalamadaka2021functional}. 

We evaluate KappaPlace across diverse VPR benchmarks, demonstrating that it provides well-calibrated uncertainty estimates while maintaining or enhancing retrieval performance. We introduce two variants: \textit{KappaPlace-JT}, a Joint-Training variant for concurrent optimization of retrieval and uncertainty estimation and \textit{KappaPlace-PT}, a Post-Training variant, where the uncertainty module is trained on top of the frozen pretrained VPR backbone. Our results show that KappaPlace consistently achieves superior Expected Calibration Error ($ECE@K$), with KappaPlace-PT reducing $ECE@1$ of state-of-the-art baselines by approximately 50\%, while maintaining high stability across random seeds. Furthermore, KappaPlace provides a principled match-level uncertainty estimate that reduces calibration error ($ECE@K$) across all benchmarks. Finally, detailed ablations justify our key design choices, namely the use of hyperspherical vMF statistics, the uncertainty head architecture and the formulation of query-level 

In summary, our contributions are as follows:
\begin{itemize}
\item We introduce a prototype-anchored supervision strategy that leverages stable latent cluster representatives as ground-truth targets, enabling aleatoric uncertainty learning while preserving VPR recall and achieving state-of-the-art $ECE@K$.
\item We design a lightweight uncertainty module that estimates the von Mises-Fisher concentration parameter ($\kappa$) in a single forward pass, supporting both joint-training and efficient post-training for frozen VPR backbones.
\item We derive a novel match-level uncertainty formulation from the $\kappa$-weighted resultant vector of query and reference embeddings, providing a principled pairwise metric to quantify the reliability of specific VPR matches.
\end{itemize}

\section{Related Work}\label{sec:related}
\subsection{Visual Place Recognition} Visual Place Recognition (VPR) identifies a query image via retrieval from a geo-tagged database. Global descriptors are typically formed using aggregation layers like NetVLAD \cite{arandjelovic2016netvlad} or GeM pooling \cite{radenovic2018fine}. Recent work enhances discriminative power through optimal transport \cite{izquierdo2024optimal}, {multi-feature and multi-view fusion} \cite{ali2023mixvpr, lu2024cricavpr}, and foundation model adaptation \cite{lu2024towards, tzachor2025effovpr}. Training objectives generally follow two paths: contrastive learning (triplet losses \cite{hoffer2015deep} and, more recently, multi-similarity \cite{Alibey_2022_gsvcities, wang2019multi} losses) which enforce condition invariance, and classification-based proxies~\cite{berton2022rethinking, berton2023eigenplaces}. Classification-based VPR scales to city-scale benchmarks by partitioning environments into discrete classes supervised by the large-margin cosine loss~\cite{berton2022rethinking, wang2018cosface}, where labels may be 'hard' or 'soft'~\cite{kim2026class}. KappaPlace leverages these classification prototypes as ground-truth mean directions to supervise our probabilistic objective.
\subsection{Uncertainty Estimation in VPR} Uncertainty is categorized as epistemic (model-based) or aleatoric (data-dependent). While MC Dropout~\cite{gal2016dropout} captures epistemic uncertainty, its multiple forward passes are computationally prohibitive for VPR. 

\textbf{Probabilistic Aleatoric Modeling.} Aleatoric methods model visual ambiguity through stochastic embeddings. The Bayesian Triplet Loss (BTL) represents images as isotropic Gaussians \cite{warburg2021bayesian} to derive uncertainty, but it is strictly coupled to triplet-loss and standard contrastive objectives. Consequently, BTL has not been demonstrated for contemporary multi-similarity losses and is inherently incompatible with scalable classification-based VPR. Furthermore, such BTL often incurs a measurable degradation in retrieval recall. STUN \cite{cai2022stun} employs teacher-student distillation to estimate uncertainty; however, it lacks an objective target and risks mimicking the latent biases of the teacher network. In contrast, we anchor uncertainty to latent prototypes (e.g., class embeddings), providing explicit supervision relative to canonical geographic representatives rather than noisy image-level estimates.

\textbf{Heuristics.} Non-probabilistic methods estimate confidence post-hoc. SUE \cite{zaffar2024estimation} measures the spatial variance of retrieved poses but remains highly sensitive to reference database density. TLD \cite{miller2026through} utilizes similarity-score statistics, such as Ratio Spread (RS), to identify ambiguous queries. Unlike these heuristics, KappaPlace provides a formal probabilistic interpretation that is independent of pose density or database distribution.
\subsection{Probabilistic Representations on the Hypersphere} VPR descriptors are typically $\ell_2$-normalized, making the von Mises-Fisher (vMF) distribution the natural choice for directional data. Its PDF aligns directly with cosine similarity, where the concentration parameter $\kappa$ captures sample-specific noise relative to a mean direction $\mu$ \cite{xie2025lh2face}. For numerical stability in high dimensions, we avoid exact modified Bessel function evaluations, and instead estimate $\kappa$ through mathematically stable bounds of the Bessel ratio \cite{amos1974computation, sra2012short}. 
\section{Methodology}\label{sec:methodology}
\subsection{Problem Formulation}
Let $\mathcal{M}$ denote the image space. A deterministic Visual Place Recognition (VPR) encoder $f_\theta: \mathcal{M} \rightarrow \mathbb{S}^{d-1}$ maps a query image $\textbf{Q} \in \mathcal{M}$ to a $d$-dimensional global descriptor $\mathbf{z} = f_\theta(Q)$, where $\|\mathbf{z}\|_2 = 1$ represents its projection onto the unit hypersphere. In classification-based VPR, the environment is partitioned into $C$ discrete geographic classes. The model is trained using a set of learnable class prototypes $\mathbf{W} = \{\mathbf{w}_1, \dots, \mathbf{w}_C\}$, where each $\mathbf{w}_j \in \mathbb{S}^{d-1}$ acts as the canonical embedding for the $j$-th location.

Standard VPR identifies the best match for a query $Q$ by finding the reference descriptor $\mathbf{z}_R$ that maximizes the cosine similarity $s(\mathbf{z}_Q, \mathbf{z}_r) = \mathbf{z}_Q^\top \mathbf{z}_R$. However, these point-estimates lack a native measure of aleatoric uncertainty; the magnitude of $s$ is uncalibrated and does not reflect the inherent noise or ambiguity present in the visual data.
\subsection{Prototype-Anchored Hyperspherical Modeling}
We propose to model image representations as random variables following a von Mises-Fisher (vMF) distribution on the unit hypersphere $\mathbb{S}^{d-1}$. The vMF probability density function for a descriptor $\mathbf{z}$ is defined as:
\begin{equation}
    p(\mathbf{z}; \boldsymbol{\mu}, \kappa) = C_d(\kappa) \exp(\kappa \boldsymbol{\mu}^\top \mathbf{z})
\end{equation}
where $\boldsymbol{\mu} \in \mathbb{S}^{d-1}$ is the mean direction and $\kappa \geq 0$ is the concentration parameter. The normalization constant $C_d(\kappa)$ is given by:
\begin{equation}
    C_d(\kappa) = \frac{\kappa^{d/2-1}}{(2\pi)^{d/2} I_{d/2-1}(\kappa)}
\end{equation}
where $I_v$ denotes the modified Bessel function of the first kind of order $v$. In this framework, $\kappa$ serves as the inverse of aleatoric uncertainty: a high $\kappa$ indicates a certain, low-noise embedding, while $\kappa \rightarrow 0$ approaches a uniform distribution on the sphere.

We anchor the vMF distribution to geographic ground truth. 
For a query image $Q$ belonging to class $j$, we set the vMF mean direction to the corresponding class prototype: $\boldsymbol{\mu}_Q = \mathbf{w}_j$. This formulation treats $\mathbf{w}_j$ as the ideal noise-free representation of a place, allowing the model to learn the concentration $\kappa_Q$ as a measure of the data-dependent deviation from this geographic anchor.

Optimization is performed by minimizing the {negative} Log-Likelihood (NLL) of the vMF distribution.  For a query $Q$, the loss $\mathcal{L}_{vMF}$ is defined as:
\begin{equation}\label{eq:vmf_loss_unstable}
    \mathcal{L}_{vMF} = \log I_{v}(\kappa_Q) -  v \log \kappa_Q - \kappa_Q \mu_Q^\top {z}_Q + \text{const}
\end{equation}
Here, $v = \frac{d}{2} - 1$, denoting the order of the modified Bessel function, which is determined by the descriptor dimensionality $d$. 

Since directly calculating the exact modified Bessel function $I_v(\kappa_Q)$ in high dimensions is numerically unstable and prone to overflow \cite{sra2012short}, we bypass the Bessel function entirely. Notice that the first two terms of Eq.~\ref{eq:vmf_loss_unstable} represent the log-partition function. Because the exact derivative of this function is the ratio of modified Bessel functions, we can substitute these first two terms with the integral of the upper bound approximation established by \cite{amos1974computation}. {By defining $\tilde{v} = v + \frac{1}{2}$ to align with the Amos bounds, we obtain the numerically stable loss:}\begin{equation}\label{eq:stable_vmf}
\mathcal{L}_{vMF} \approx \left( \sqrt{\kappa_Q^2 + \tilde{v}^2} - \tilde{v} \log\left(\tilde{v} + \sqrt{\kappa_Q^2 + \tilde{v}^2}\right) \right) - \kappa_Q \mu_Q^\top z_Q + \text{const}\end{equation}

While we focus on an anchoring paradigm for classification-based VPR, the supervision principle of KappaPlace can be applied to contrastive frameworks.  In the absence of class prototypes, we utilize Batch-Centroid Anchoring. For a query $Q$ with a set of $n$ known positive samples $\{P_1, \dots, P_n\}$ within a training batch, we compute the positive centroid as the temporary anchor:
\begin{equation}
\mathbf{\bar{w}}_Q = \frac{\sum_{j=1}^{n} \mathbf{z}_{P_j}}{|\sum_{j=1}^{n} \mathbf{z}_{P_j}|_2}
\end{equation}
We then set the vMF mean direction to this anchor: $\boldsymbol{\mu}_Q = \mathbf{\bar{w}}_Q$ and optimize with the vMF loss. This allows the model to quantify uncertainty relative to the local visual consensus of a location within a contrastive training paradigm.

\subsection{Estimating the Concentration Parameter ($\kappa$)}  To estimate $\kappa$, we introduce a lightweight regressor $g_\phi$ that operates in parallel with the standard retrieval path. For a query image $Q$, the frozen backbone $f_\theta$ generates a latent feature map. While the retrieval path produces a deterministic global descriptor $\mathbf{z}_Q \in \mathbb{S}^{d-1}$, the Kappa Head maps the same features to a scalar concentration parameter $\kappa_Q > 0$. The design of $g_\phi$ mirrors the aggregation layers of the underlying VPR backbone to ensure that uncertainty is estimated at the same level of feature abstraction as the global descriptor. Following this  aggregation, $g_\phi$ employs a final linear projection and a Softplus activation to estimate $\kappa$. This activation guarantees the strictly positive values required for a valid von Mises-Fisher distribution.

\subsection{Training Paradigms}
\paragraph{Post Training.}{In this variant, the base architecture is first trained to get the class weights. Subsequently,} the VPR backbone $f_\theta$ {and these weights are} kept \textit{frozen}, and only the parameters $\phi$ of the Kappa Head are optimized. This mode is computationally efficient and ideal for adding uncertainty estimation to existing, high-performance models without altering their established discriminative features. 

The objective is simply the minimization of the vMF loss defined in Eq.~\ref{eq:stable_vmf}:
\begin{equation}\label{eq:total_frozen_loss}
\mathcal{L}_{total} = \mathcal{L}_{vMF}({z}_Q, \mu_Q, \kappa_Q)
\end{equation}
where $\mathbf{z}_Q$ is the fixed descriptor and $\mu_Q$ is the fixed class prototype for the class that sample $Q$ belongs to.

\paragraph{Joint Training.} While the Kappa Head $g_\phi$ can be trained as a post-hoc reliability layer for a frozen backbone, the framework natively supports Joint Training. In this mode, both the backbone encoder $f_\theta$, the prototype matrix $W$ and the uncertainty head $g_\phi$ are jointly optimized. 

The total objective is a multi-task loss:
\begin{equation}\label{eq:total_joint_loss}
\mathcal{L}_{total} = \mathcal{L}_{cls} + \lambda \mathcal{L}_{vMF}
\end{equation}
where $\mathcal{L}_{cls}$ is the Large Margin Cosine Loss, used to train classification-based VPR methods~\cite{berton2022rethinking} and $\lambda$ is a balancing hyperparameter. Joint training allows the backbone to learn "uncertainty-aware" features, where the latent space is structured to explicitly separate distinctive landmarks from ambiguous, high-noise environmental elements.

\subsection{Query-level and Match-level Uncertainty Estimation}\label{subsec:unc_estimation}
\paragraph{Query-level Uncertainty Estimation.} Given a query image $Q$, we compute its global descriptor ${z}_Q$ and estimate its concentration $\kappa_Q$ through a single forward pass. While a straightforward measure of query uncertainty would be the inverse concentration $1/\kappa_Q$, we propose a more holistic uncertainty score that accounts for the relationship between the query and its retrieved context. 

Specifically, we consider both $\kappa_Q$ and $\kappa_{R_1}$, the concentration parameter of the top-ranked neighbor $R_1$ to quantify the reliability of the retrieval. Drawing from the properties of the von Mises-Fisher (vMF) distribution, we define $U_Q$, the uncertainty score for $Q$, as the inverse magnitude of the resultant vector formed by the vector summation of the query and retrieval parameters~\cite{jammalamadaka2021functional}. Intuitively, this represents the (inverse of the) concentration of the fused distributions:
\begin{equation}\label{eq:query_uncertainty}
U_Q= \frac{1}{\sqrt{\kappa_Q^2 + \kappa_{R_1}^2 + 2\kappa_Q\kappa_{R_1}({z}_Q^\top{z}_{R_1})}}
\end{equation}
where ${z}_{R_1}$ is the global descriptor of $R_1$.
In our ablation study (Tables\ref{tab:unc_score_city}-\ref{tab:unc_score_environmental} in the Appendix), we show that this formulation improves calibration compared to using $\kappa_Q$ alone {in datasets featuring severe domain shifts, such as Amstertime and MSLS-val.}

\paragraph{Match-level Uncertainty Estimation.}
Existing VPR methods primarily focus on query-level reliability (i.e., "is this image difficult to localize?"). However, in downstream robotics and SLAM pipelines, the reliability of a \textit{specific} match is often more critical. For any query-reference pair $(Q, R)$, we define the match-level uncertainty $U_{Q \leftrightarrow R}$ by substituting their respective concentration parameters into the resultant vector formula (Eq.~\ref{eq:query_uncertainty}):
\begin{equation}\label{eq:match_uncertainty}
U_{Q \leftrightarrow R} = \frac{1}{\sqrt{\kappa_Q^2 + \kappa_{R}^2 + 2\kappa_Q\kappa_{R}({z}_Q^\top {z}_{R})}}
\end{equation}

This formulation allows the system to distinguish between a high-confidence match (where both descriptors are certain and aligned) and an accidental alignment of two uncertain descriptors. By rewarding "accumulated evidence", our approach ensures that two confident descriptors pointing in the same direction yield a lower uncertainty score than two uncertain ones, even if their angular similarity is identical.

\subsection{Evaluating Uncertainty in Retrieval and Matching}~\label{subsec:unc_evaluation}A well-calibrated uncertainty estimate accurately reflects the probability of success. In VPR, success is measured by Recall@$K$ ($R@K$), the event that at least one of the top $K$ retrieved neighbors falls within a ground-truth distance threshold (e.g., 25m). We utilize the Expected Calibration Error (ECE) to measure the alignment between predicted uncertainties and actual recognition performance (success). Following~\cite{warburg2021bayesian}, we compute $ECE@K$ by partitioning $N$ test queries into $M$ bins sorted by their predicted uncertainty $U_Q$ (Eq.~\ref{eq:query_uncertainty}). We assign each bin an expected success level $\mathcal{C}$ based on its rank-order, where the most certain bin is expected to achieve perfect recall ($\mathcal{C}=1.0$) and the least certain bin is expected to have the lowest recall ($\mathcal{C}=0.0$). The calibration error is defined as the discrepancy between this expectation and the observed recall:
\begin{equation}\label{eq:ECE_K}ECE@K = \sum_{i=1}^{M} \frac{|B_i|}{N} \left| R@K(B_i) - \mathcal{C}(B_i) \right|
\end{equation}
where $R@K(B_i)$ is the actual $R@K$ achieved by the queries within bin $B_i$. We also leverage the ECE@K metric to evaluate match-level uncertainty as explained in Appendix~\ref{subsec:appendix_ece}.

\section{Experimental Results}
\label{sec:results}

\subsection{Experimental Setup}\label{sec:impl}
\paragraph{Datasets.} We evaluate KappaPlace across a diverse suite of VPR benchmarks featuring challenging viewpoint changes, environmental shifts, and perceptual aliasing. Our city-scale evaluation utilizes Pittsburgh 30k (Pitts30k)~\cite{pitts} and the test sets from the San Francisco XL (SF-XL) dataset~\cite{berton2022rethinking}. To assess resilience to environmental and long-term temporal variations, we also evaluate performance on the Mapillary Street Level Sequences (MSLS-val)~\cite{msls} and AmsterTime~\cite{amstertime} datasets.

\paragraph{Baselines.} We use a ResNet-50 CosPlace~\cite{berton2022rethinking} backbone for all methods, comparing two categories of approaches:
\begin{itemize}
\item \textit{Post-Training:} These methods utilize a frozen backbone to estimate uncertainty post-hoc, preserving the original $Recall@K$. We compare \textit{L2-distance} (top-match distance), \textit{PA-score} (ratio of two nearest neighbors), \textit{SUE}~\cite{zaffar2024estimation} (spatial Gaussian fit over $K$-best poses), and our proposed \textit{KappaPlace-PT}.
\item \textit{Joint-Training:} These methods concurrently optimize retrieval and uncertainty tasks. We compare our \textit{KappaPlace-JT} against \textit{STUN}~\cite{cai2022stun}, which we re-implemented using the CosPlace architecture for both teacher and student models to ensure a fair comparison.
\end{itemize}

\paragraph{Evaluation Protocol.} Retrieval performance is measured via $Recall@K$ with a standard 25-meter ground-truth threshold. To assess the reliability of query-level and match-level uncertainty, we utilize $ECE@K$, as detailed in Section~\ref{subsec:unc_evaluation} 

\paragraph{Implementation Details.}
We instantiate KappaPlace using a ResNet-50 backbone for main results and ResNet-18 for ablation studies. The uncertainty head $g_\phi$ estimates the concentration $\kappa$ by mirroring the CosPlace aggregation module which applies $L_2$-normalization, GeM-pooling, flattening and a linear transformation. This is followed by a Linear layer to map the flattened features to a single $\kappa$ value and Softplus to ensure a non-negativity. Additional training and implementation details are provided in Appendix~\ref{sec:appendix_impl_details}.

\subsection{Query-level Uncertainty Estimation}
\begin{figure*}[tbh!]
    \centering
    \begin{tikzpicture}
        \begin{axis}[
            ybar,
            bar width=4.1pt,          
            width=0.7\textwidth,    
            height=4cm,           
            ylabel={ECE@K $\downarrow$},
            ylabel style={font=\footnotesize, yshift=-5pt},
            symbolic x coords={K=1, K=5, K=10},
            xtick=data,
            ymin=0, ymax=1.15,      
            enlarge x limits=0.3,
            legend style={
                at={(0.5,-0.2)},    
                anchor=north, 
                legend columns=-1,   
                font=\tiny, 
                draw=none,          
                column sep=6pt      
            },
            grid=major,
            grid style={dashed, gray!20},
            nodes near coords,
            every node near coord/.append style={
                rotate=90, 
                anchor=west, 
                font=\tiny, 
                /pgf/number format/fixed, 
                /pgf/number format/precision=3
            },
            tick label style={font=\footnotesize},
            label style={font=\footnotesize}
        ]
            \addplot[fill=gray!30, draw=gray!60] coordinates {(K=1,0.492) (K=5,0.554) (K=10,0.569)};
            
            \addplot[fill=gray!60, draw=gray!80] coordinates {(K=1,0.737) (K=5,0.799) (K=10,0.814)};
            
            \addplot[fill=cyan!40, draw=cyan!70] coordinates {(K=1,0.135) (K=5,0.191) (K=10,0.204)};
            
            \addplot[fill=red!40, draw=red!70] coordinates {(K=1,0.378) (K=5,0.436) (K=10,0.450)};
            
            \addplot[fill=green!40, draw=green!70] coordinates {(K=1,0.066) (K=5,0.117) (K=10,0.129)};
            
            \addplot[fill=green!70, draw=green!90] coordinates {(K=1,0.104) (K=5,0.154) (K=10,0.165)};

            \legend{L2, PA, SUE, STUN, \textbf{KP-PT}, \textbf{KP-JT}}
        \end{axis}
    \end{tikzpicture}
    \caption{Aggregated ECE@K calibration results across  benchmarks. 
    }
    \label{fig:combined_ece_final}
\end{figure*}
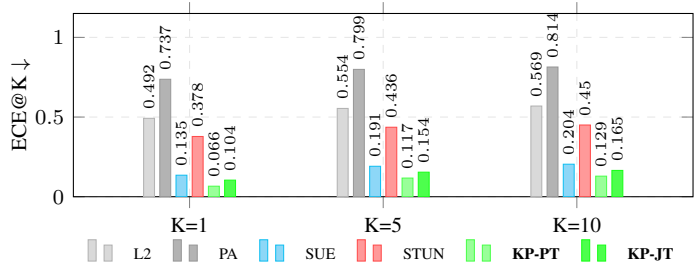

\begin{figure*}[tbh]
    \centering
    \begin{subfigure}[b]{0.32\textwidth}
        \centering
        \includegraphics[width=\textwidth]{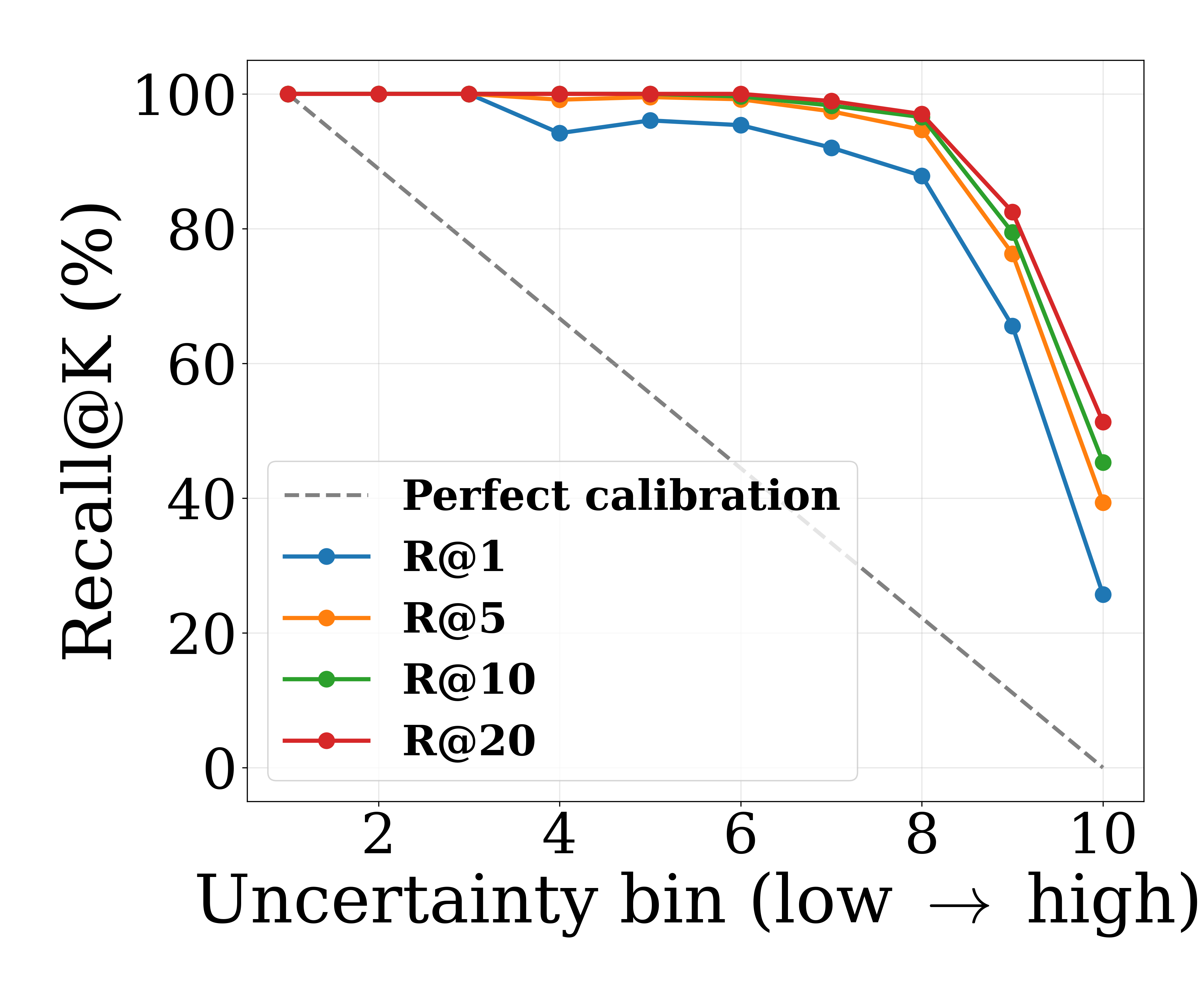}
        \caption{L2 (Baseline)}
        \label{fig:recall_bin_l2}
    \end{subfigure}
    \hfill
    \begin{subfigure}[b]{0.32\textwidth}
        \centering
        \includegraphics[width=\textwidth]{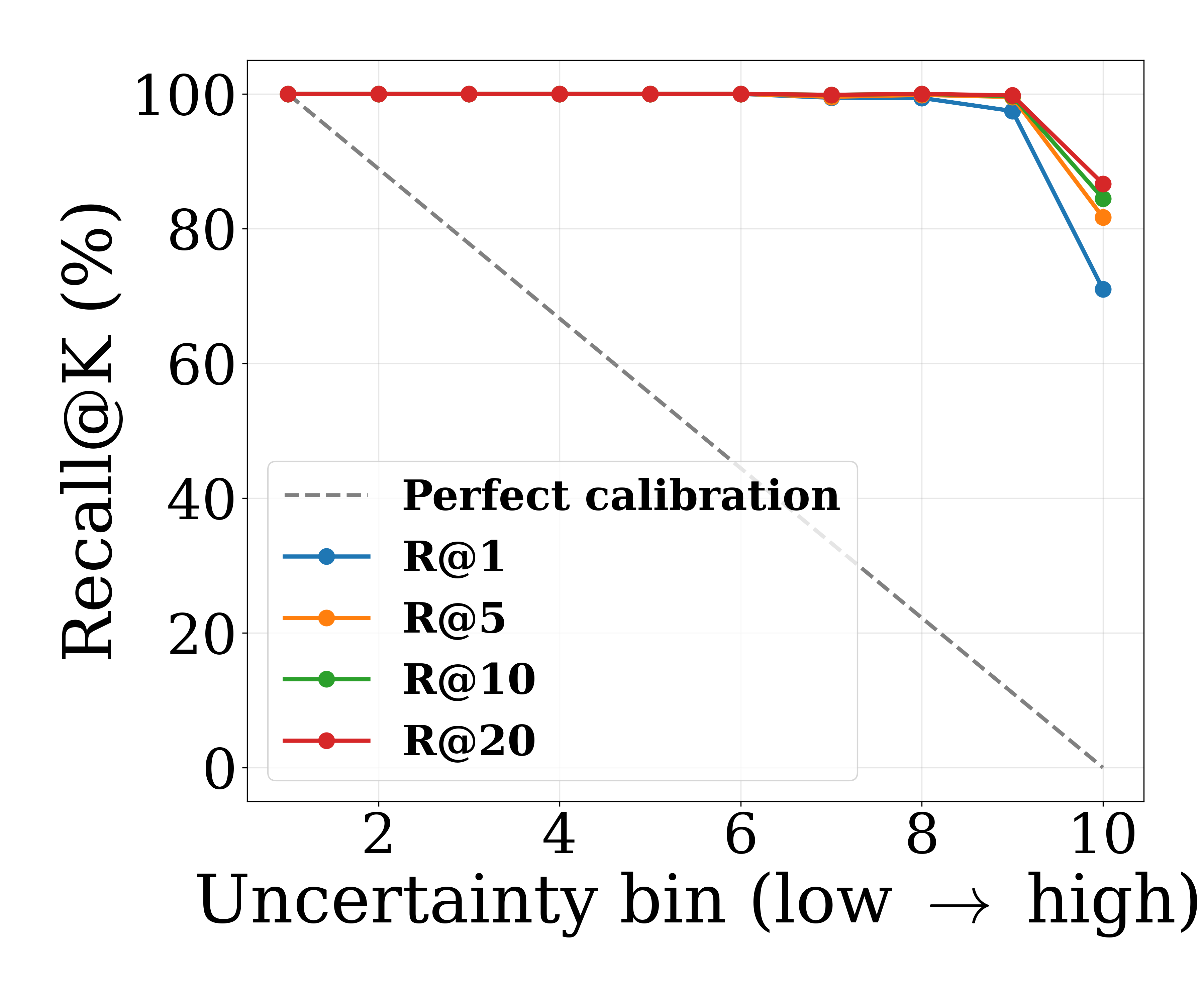}
        \caption{PA-Score}
        \label{fig:recall_bin_pa}
    \end{subfigure}
    \hfill
    \begin{subfigure}[b]{0.32\textwidth}
        \centering
        \includegraphics[width=\textwidth]{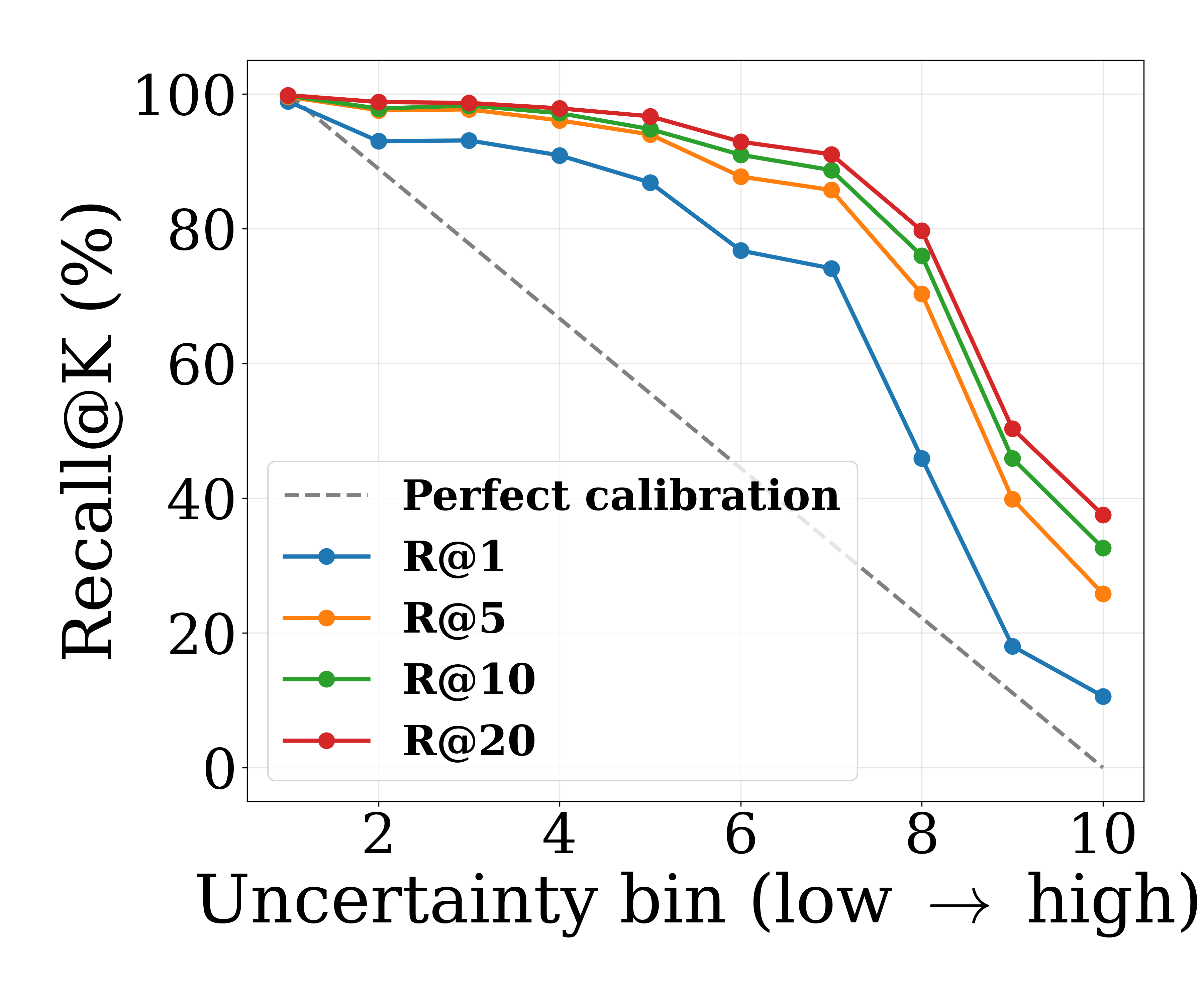}
        \caption{SUE}
        \label{fig:recall_bin_sue}
    \end{subfigure}

    \vspace{1em} 

    \begin{subfigure}[b]{0.32\textwidth}
        \centering
        \includegraphics[width=\textwidth]{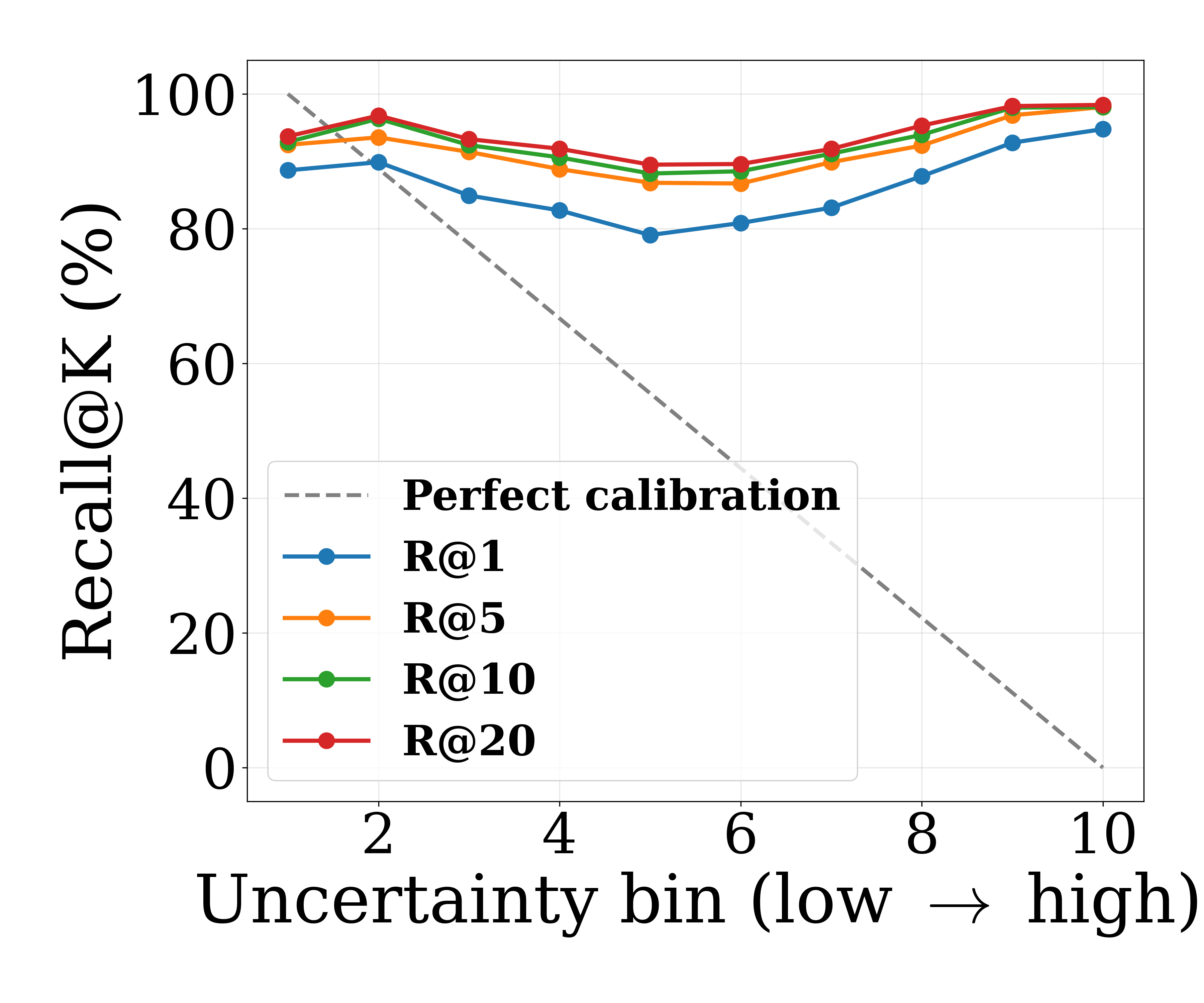}
        \caption{STUN}
        \label{fig:recall_bin_stun}
    \end{subfigure}
    \hfill
    \begin{subfigure}[b]{0.32\textwidth}
        \centering
        \includegraphics[width=\textwidth]{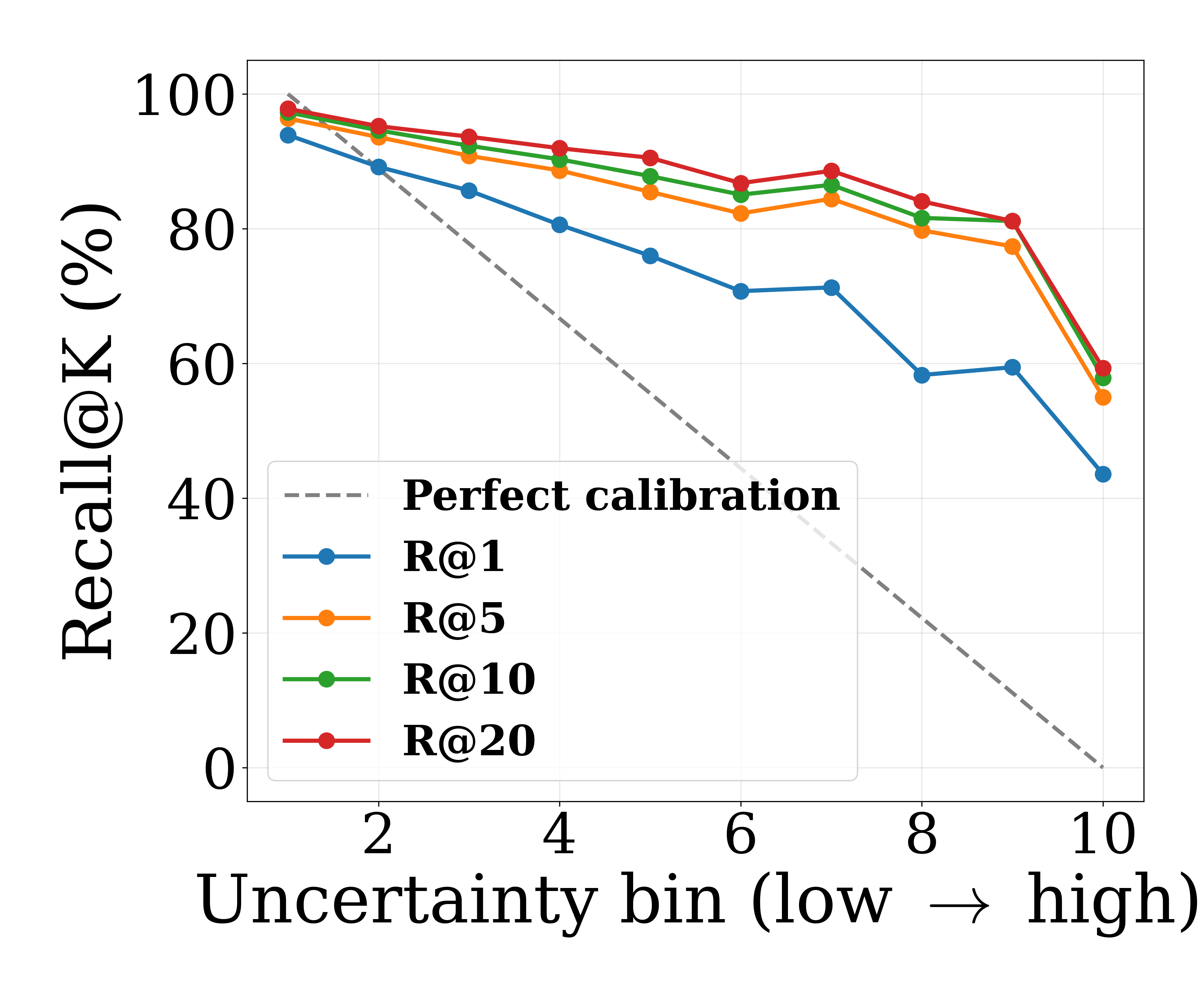}
        \caption{KappaPlace-JT}
        \label{fig:recall_bin_jt}
    \end{subfigure}
    \hfill
    \begin{subfigure}[b]{0.32\textwidth}
        \centering
        \includegraphics[width=\textwidth]{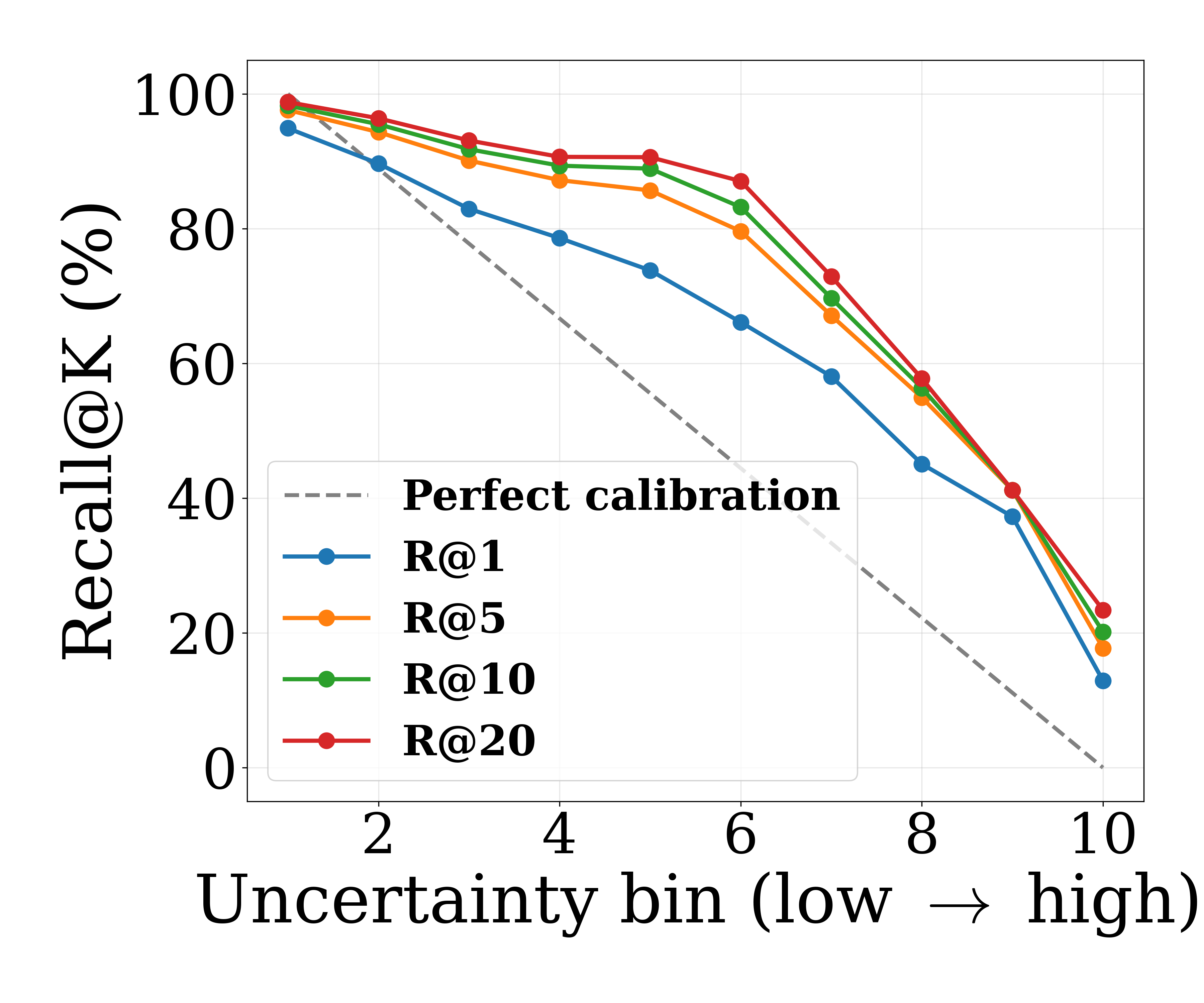}
        \caption{KappaPlace-PT}
        \label{fig:recall_bin_pt}
    \end{subfigure}

    \caption{Aggregated Recall@K results across uncertainty-sorted bins. The diagonal represents perfect calibration, where the predicted uncertainty bin precisely matches the empirical success rate.} 
    \label{fig:recall_uncertainty_analysis}
\end{figure*}

Tables \ref{tab:sf_final} and \ref{tab:ams_msls_final} summarize retrieval and calibration performance across diverse urban and long-term benchmarks with a ResNet-50 CosPlace backbone. KappaPlace variants consistently achieve the lowest Expected Calibration Error ($ECE@K$), outperforming both post-hoc and joint-training baselines in nearly every scenario. As shown in the aggregated results (Fig.~\ref{fig:combined_ece_final}), both variants substantially outperform SUE, with KappaPlace-PT reducing SUE's calibration error by approximately 50\% at $K=1$. The ECE calibration curves in Fig.~\ref{fig:recall_uncertainty_analysis} further validate these results. While both KappaPlace and SUE align closely with the "perfect calibration" diagonal, KappaPlace-PT demonstrates the most consistent and well-calibrated performance across all uncertainty bins. 
Crucially, while dataset pooling can mask local calibration failures via error cancellation, the analysis in Fig.~\ref{fig:recall_uncertainty_analysis} shows that KappaPlace-PT maintains a strong, monotonic alignment with the diagonal across the entire uncertainty spectrum. This confirms that our low aggregated ECE stems from robust, domain-agnostic mapping rather than fortuitous cancellation. Furthermore, whereas baseline calibration often deteriorates under dataset pooling, our method remains stable, indicating a reliable signal for diverse real-world deployment.

We note that methods relying on neighbor consistency (like SUE) or those requiring ground-truth poses for geometric verification are unsuitable for datasets such as AmsterTime, which contain only a single true positive per query and lack 6DoF pose labels. In contrast, KappaPlace provides reliable uncertainty estimates in a single forward pass, independent of neighbor density or pose metadata. To avoid a negative bias against SUE and ensure a fair comparison, we omit AmsterTime from the aggregated results in Figs.~\ref{fig:combined_ece_final}--\ref{fig:recall_uncertainty_analysis}.

Our two variants offer a flexible deployment strategy: KappaPlace-PT provides state-of-the-art calibration while strictly preserving the recall of frozen backbones, whereas KappaPlace-JT also enhances accuracy. For instance, on SF-Test-v1, KappaPlace-JT improves Recall@1 by nearly 3\% while reducing calibration error by 77\% relative to the $L_2$ baseline. This allows practitioners to balance computational budgets with performance requirements. In contrast, while STUN improves calibration over $L_2$, it frequently degrades retrieval recall (red highlights in Tables \ref{tab:sf_final}--\ref{tab:ams_msls_final}). We attribute this to STUN’s distillation against pointwise embeddings, which fails to capture the underlying manifold as effectively as our prototype-anchoring strategy.

\begin{table*}[t]
\centering
\small
\caption{Retrieval and calibration results on San Francisco benchmarks.}
\label{tab:sf_final}
\resizebox{\textwidth}{!}{
\begin{tabular}{l ccc ccc ccc ccc}
\toprule
\multirow{2}{*}{\textbf{Method}} & \multicolumn{6}{c}{\textbf{SF-Test-v1}} & \multicolumn{6}{c}{\textbf{SF-Test-v2}} \\
\cmidrule(lr){2-7} \cmidrule(lr){8-13}
 & \multicolumn{3}{c}{Recall@K $\uparrow$} & \multicolumn{3}{c}{ECE@K $\downarrow$} & \multicolumn{3}{c}{Recall@K $\uparrow$} & \multicolumn{3}{c}{ECE@K $\downarrow$} \\
 & 1 & 5 & 10 & 1 & 5 & 10 & 1 & 5 & 10 & 1 & 5 & 10 \\
\midrule
\rowcolor{gray!10} \multicolumn{13}{l}{\textit{Post-Training Methods (Frozen Backbone)}} \\
L2 (Baseline) & \underline{76.0} & 83.2 & 85.1 & 0.421 & 0.493 & 0.512 & \underline{88.5} & \underline{94.6} & \underline{96.7} & 0.391 & 0.453 & 0.473 \\
PA-Score & \multicolumn{3}{c}{\multirow{3}{*}{--- \textit{Preserves L2} ---}} & 0.641 & 0.713 & 0.732 & \multicolumn{3}{c}{\multirow{3}{*}{--- \textit{Preserves L2} ---}} & 0.742 & 0.804 & 0.824 \\
SUE & \multicolumn{3}{c}{} & 0.207 & 0.279 & 0.298 & \multicolumn{3}{c}{} & 0.275 & 0.331 & 0.351 \\
\textbf{KappaPlace-PT} & \multicolumn{3}{c}{} & \textbf{0.093} & \underline{0.163} & \underline{0.182} & \multicolumn{3}{c}{} & \textbf{0.172} & \textbf{0.231} & \textbf{0.251} \\
\midrule
\rowcolor{gray!10} \multicolumn{13}{l}{\textit{Joint-Training Methods (Fine-tuned Backbone)}} \\
STUN & \textcolor{red}{75.6} & \underline{84.1} & \underline{86.1} & 0.288 & 0.355 & 0.371 & \textcolor{red}{86.8} & \textcolor{red}{93.5} & \textcolor{red}{95.3} & 0.350 & 0.417 & 0.432 \\
\textbf{KappaPlace-JT} & \textbf{78.9} & \textbf{85.5} & \textbf{87.9} & \underline{0.097} & \textbf{0.154} & \textbf{0.174} & \textbf{91.3} & \textbf{95.8} & \textbf{97.0} & \underline{0.220} & \underline{0.262} & \underline{0.270} \\
\bottomrule
\end{tabular}}
\end{table*}
\begin{table*}[tbh!]
\centering
\small
\caption{Retrieval and calibration results on AmsterTime and MSLS-val benchmarks.}
\label{tab:ams_msls_final}
\resizebox{\textwidth}{!}{
\begin{tabular}{l ccc ccc ccc ccc}
\toprule
\multirow{2}{*}{\textbf{Method}} & \multicolumn{6}{c}{\textbf{AmsterTime}} & \multicolumn{6}{c}{\textbf{MSLS-val}} \\
\cmidrule(lr){2-7} \cmidrule(lr){8-13}
 & \multicolumn{3}{c}{Recall@K $\uparrow$} & \multicolumn{3}{c}{ECE@K $\downarrow$} & \multicolumn{3}{c}{Recall@K $\uparrow$} & \multicolumn{3}{c}{ECE@K $\downarrow$} \\
 & 1 & 5 & 10 & 1 & 5 & 10 & 1 & 5 & 10 & 1 & 5 & 10 \\
\midrule
\rowcolor{gray!10} \multicolumn{13}{l}{\textit{Post-Training Methods (Frozen Backbone)}} \\
L2 (Baseline) & \underline{45.7} & \underline{66.0} & \underline{72.9} & \underline{0.209} & 0.412 & 0.480 & \textbf{87.0} & \underline{92.0} & \underline{94.2} & 0.380 & 0.430 & 0.452 \\
PA-Score & \multicolumn{3}{c}{\multirow{3}{*}{--- \textit{Preserves L2} ---}} & 0.374 & 0.578 & 0.646 & \multicolumn{3}{c}{\multirow{3}{*}{--- \textit{Preserves L2} ---}} & 0.716 & 0.766 & 0.788 \\
SUE & \multicolumn{3}{c}{} & --- & --- & --- & \multicolumn{3}{c}{} & 0.209 & 0.243 & 0.264 \\
\textbf{KappaPlace-PT} & \multicolumn{3}{c}{} & 0.310 & \textbf{0.145} & \textbf{0.108} & \multicolumn{3}{c}{} & \underline{0.074} & \underline{0.116} & \underline{0.138} \\
\midrule
\rowcolor{gray!10} \multicolumn{13}{l}{\textit{Joint-Training Methods (Fine-tuned Backbone)}} \\
STUN & \textcolor{red}{43.7} & \textcolor{red}{64.7} & \textcolor{red}{71.6} & \textbf{0.191} & 0.222 & 0.257 & \textcolor{red}{84.7} & \textcolor{red}{90.9} & \textcolor{red}{92.3} & 0.344 & 0.395 & 0.409 \\
\textbf{KappaPlace-JT} & \textbf{47.5} & \textbf{68.0} & \textbf{74.0} & 0.339 & \underline{0.172} & \underline{0.133} & \textcolor{red}{\underline{86.6}} & \textbf{93.0} & \textbf{94.3} & \textbf{0.040} & \textbf{0.071} & \textbf{0.076} \\
\bottomrule
\end{tabular}}
\end{table*}

\subsection{Match-level Uncertainty Estimation}
\begin{figure}[h!]\centering\includegraphics[scale=0.4]{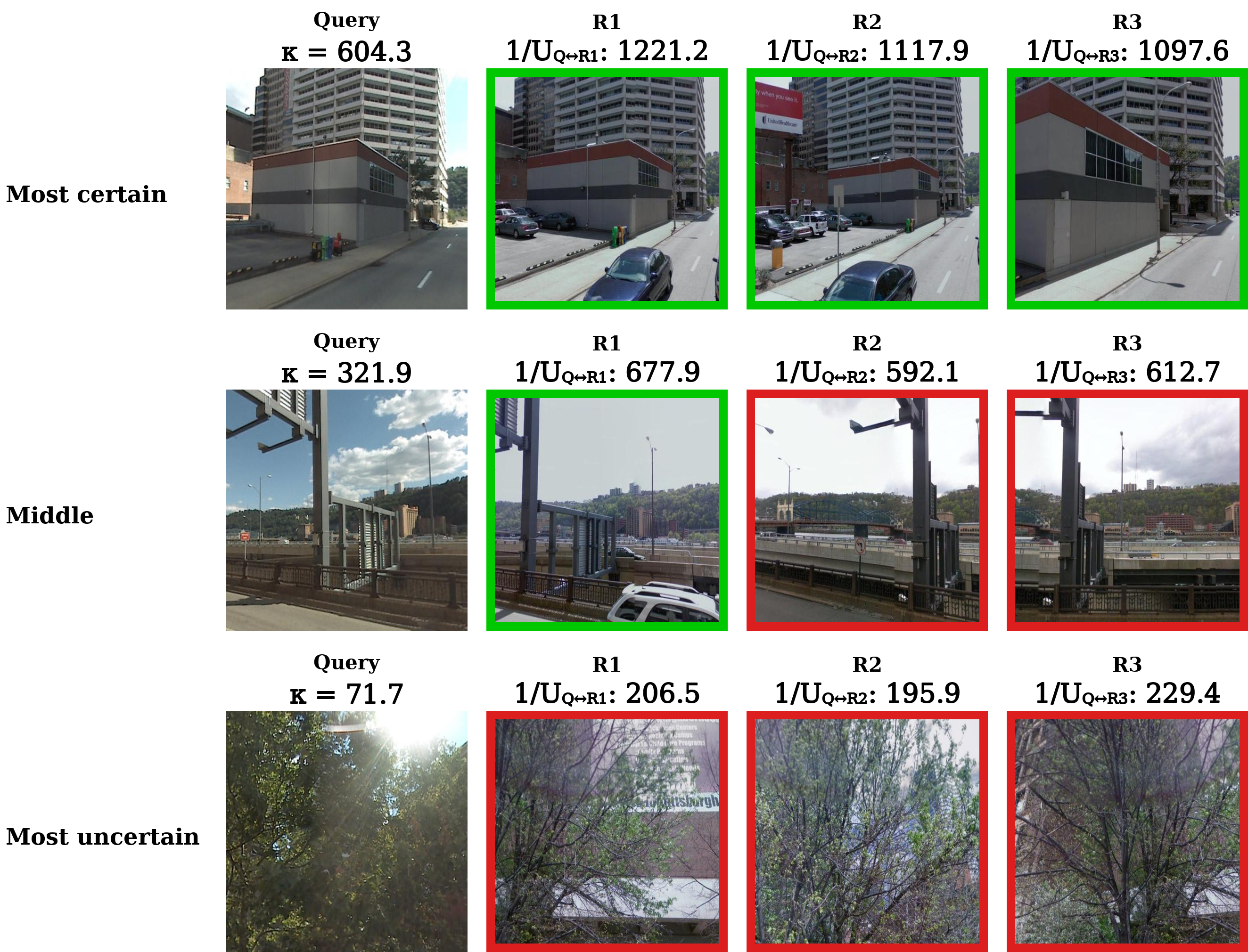}
\caption{Match-level Confidence Analysis. Top-3 retrievals for queries with high, medium, and low reliability, indicated by their $\kappa$ parameter value. Confidence scores per match, shown on top of each retrieved reference image, are derived from the match-level uncertainty of (Eq.~\ref{eq:match_uncertainty}), with green/red borders indicating correct/incorrect matches. 
}
\label{fig:reterival_match_examples}
\end{figure}

Table \ref{tab:match_level_selected} summarizes calibration results for match-level uncertainty estimation. While most VPR research focuses on query-level difficulty, our approach provides a principled mechanism to evaluate the reliability of specific query-reference pairs. By applying our resultant vector formulation (Eq. \ref{eq:match_uncertainty}), KappaPlace consistently reduces match-level $ECE@K$ compared to the $L_2$ baseline across the majority of benchmarks. Notably, on 
{MSLS-val}, KappaPlace-JT reduces calibration error by nearly an order of magnitude at lower $K$ values. We visualize these match-level confidence scores in Fig.~\ref{fig:reterival_match_examples}. For three representative queries, we display the top-3 references ($R_1$–$R_3$) alongside their confidence estimates (the reciprocal of the match-level uncertainty). The scores effectively rank retrievals by reliability: "High Confidence" cases yield high concentration for all correct matches, while "Mid-range" scores correctly distinguish valid matches from incorrect retrievals caused by perceptual similarity. Finally, "Low Confidence" queries, dominated by non-discriminative foliage, receive substantially lower scores, correctly signaling that all candidates are likely incorrect.
\begin{table*}[tbh!]
\centering
\small
\caption{Match-level ECE@K Results on urban and long-term benchmarks.}
\label{tab:match_level_selected}
\resizebox{\textwidth}{!}{
\begin{tabular}{lcccccccccccc}
\toprule
\textbf{Method} & \multicolumn{3}{c}{\textbf{SF-Test-v1}} & \multicolumn{3}{c}{\textbf{SF-Test-v2}} & \multicolumn{3}{c}{\textbf{Pitts30k}} & \multicolumn{3}{c}{\textbf{MSLS-val}} \\
\cmidrule(lr){2-4} \cmidrule(lr){5-7} \cmidrule(lr){8-10} \cmidrule(lr){11-13}
  & \multicolumn{3}{c}{ECE@K $\downarrow$} & \multicolumn{3}{c}{ECE@K $\downarrow$} & \multicolumn{3}{c}{ECE@K $\downarrow$} & \multicolumn{3}{c}{ECE@K $\downarrow$} \\
  & 1 & 5 & 10 & 1 & 5 & 10 & 1 & 5 & 10 & 1 & 5 & 10 \\
\midrule
L2 & 0.421 & 0.359 & 0.275 & 0.391 & 0.418 & 0.392 & 0.516 & 0.482 & 0.398 & 0.380 & 0.315 & \textbf{0.189} \\
\midrule
{KappaPlace-PT} & \textbf{0.093} & 0.112 & 0.243 & \textbf{0.172} & \textbf{0.133} & \textbf{0.095} & 0.191 & 0.094 & \textbf{0.129} & 0.074 & \textbf{0.118} & 0.292 \\
{KappaPlace-JT} & 0.097 & \textbf{0.078} & \textbf{0.210} & 0.220 & 0.155 & 0.104 & \textbf{0.175} & \textbf{0.088} & 0.146 & \textbf{0.040} & 0.187 & 0.363 \\
\bottomrule
\end{tabular}}
\end{table*}

\subsection{Ablations \& Stability Analysis}
Table \ref{tab:ablations_combined} summarizes our ablation study regarding the loss function and uncertainty head architecture, conducted with a ResNet18 CosPlace encoder. We compare our vMF-based objective against a Gaussian Negative Log-Likelihood (GNLL) variant and evaluate the impact of aggregation layers versus a simple linear projection for uncertainty estimation. 
While GNLL is a standard choice for regression-based uncertainty, its performance is inferior in this context, likely because the vMF distribution naturally aligns with the hyperspherical geometry of VPR embeddings. 
In addition, we find that mirroring the backbone aggregation provides the necessary capacity to map descriptor characteristics to the concentration parameter $\kappa$, as evidenced by the performance drop when using a a single-layer linear head.

To evaluate robustness, Table \ref{tab:seed_results_vertical} presents calibration performance across five training runs with different random seeds. KappaPlace demonstrates high stability, with standard deviations consistently low ($\leq 0.01$) across all benchmarks and $K$ depths. This stability suggests our prototype-anchored objective provides a reliable signal for uncertainty estimation that is robust to initialization. Such consistency is vital for safety-critical robotics, where predictable behavior across deployments is essential. Finally, we report inference latency, memory overhead, training time, and compute details for KappaPlace in the Appendix. 

\begin{table*}[tbh!]
\centering
\small
\caption{Ablation study results aggregated across all benchmarks. We compare different loss functions and uncertainty head architectures to justify our final model configuration.}
\label{tab:ablations_combined}
\resizebox{0.7\textwidth}{!}{
\begin{tabular}{lllccc}
\toprule
\textbf{Config. Name} & \textbf{Loss} & \textbf{Head} & \multicolumn{3}{c}{ECE@K $\downarrow$} \\
\cmidrule(lr){4-6}
 & &  & 1 & 5 & 10 \\
\midrule
GNLL-variant & GNLL & Agg. + Linear + Softplus & 0.230 & 0.322 & 0.350 \\
Linear-head & vMF & Linear + Softplus & 0.179 & 0.267 & 0.295 \\
{KappaPlace-PT (Ours)} & vMF & Agg. + Linear + Softplus  & \textbf{0.080} & \textbf{0.172} & \textbf{0.200} \\
\bottomrule
\end{tabular}}
\end{table*}

\begin{table*}[tbh!]
\centering
\small
\caption{ECE@K for KappaPlace-PT averaged over multiple seeds (mean $\pm$ std).}
\label{tab:seed_results_vertical}
\setlength{\tabcolsep}{10pt} 
\resizebox{0.6\textwidth}{!}{
\begin{tabular}{lccc}
\toprule
\textbf{Dataset} & \textbf{ECE@1} & \textbf{ECE@5} & \textbf{ECE@10} \\
\midrule
SF-Test-v1 & 0.212 $\pm$ 0.007 & 0.112 $\pm$ 0.006 & 0.085 $\pm$ 0.007 \\
SF-Test-v2 & 0.157 $\pm$ 0.009 & 0.261 $\pm$ 0.009 & 0.287 $\pm$ 0.009 \\
\midrule
Combined & 0.079 $\pm$ 0.003 & 0.170 $\pm$ 0.004 & 0.198 $\pm$ 0.004 \\
\bottomrule
\end{tabular}}
\end{table*}

\section{Conclusion}
\textbf{Limitations and Future Work.} While KappaPlace achieves state-of-the-art calibration, evaluation in retrieval remains underdeveloped compared to classification. Specifically, $ECE@K$ is subject to binning sensitivities and does not fully capture ranking reliability, highlighting a critical need for new, retrieval-specific uncertainty evaluation metrics. In addition, while we provide formulation for both classification and contrastive-based approaches, our study focused on the uncertainty estimation for classification-based VPR which is relatively under-explored. Future work will extend this evaluation to contrastive methods.

\textbf{Summary.} In this work, we introduced KappaPlace, a novel Prototype-Anchored supervision strategy for VPR which enables query-level and match-level uncertainty estimation without sacrificing performance. Beyond the reported improvements, KappaPlace enables VPR to serve as a reliable component for safety-critical robotics and autonomous navigation. 

\bibliography{uvpr}
\bibliographystyle{plain}

\appendix

\section{Appendix}
We provide additional details regarding our experimental setup, as well as extended  results that supplement the findings in the main text.

\subsection{Per-Dataset Retrieval and Calibration Performance}
Table~\ref{tab:pitts_appendix} provides the detailed Recall@K and ECE@K breakdown for the Pitts30k dataset.
\subsection{Extended Ablation Results (Per-Dataset Analysis)}
Tables~\ref{tab:ablations_city_scale} and \ref{tab:ablations_environmental} present the per-dataset breakdown of the ablations reported in the main text. We evaluate the sensitivity of benchmarks to the loss formulation (vMF vs. GNLL) and the uncertainty head architecture (mirroring the aggregation layers of the backbone + Linear + Softplus (Agg.) vs. Linear + Softplus). The results demonstrate that the combination of vMF-based supervision and an aggregation-based head yields superior robustness across multiple benchmarks.

\subsection{Ablation of Query-level Uncertainty Formulations} 
We compare different formulations for the final query-level uncertainty score in Tables~\ref{tab:unc_score_city} and \ref{tab:unc_score_environmental}. Our results justify the use of the resultant vector magnitude approach over the naive inverse-kappa ($1/\kappa_Q$) baseline, particularly in environments prone to perceptual aliasing.
\begin{table*}[tbh!]
\centering
\small
\caption{Retrieval and Calibration results on the Pitts30k Benchmark.}
\label{tab:pitts_appendix}
\resizebox{0.7\textwidth}{!}{
\begin{tabular}{l ccc ccc}
\toprule
\multirow{2}{*}{\textbf{Method}} & \multicolumn{3}{c}{\textbf{Recall@K} $\uparrow$} & \multicolumn{3}{c}{\textbf{ECE@K} $\downarrow$} \\
\cmidrule(lr){2-4} \cmidrule(lr){5-7}
 & 1 & 5 & 10 & 1 & 5 & 10 \\
\midrule
\rowcolor{gray!10} \multicolumn{7}{l}{\textit{Post-Training Methods (Frozen Backbone)}} \\
L2 (Baseline) & \underline{89.4} & \textbf{95.3} & \textbf{96.6} & 0.516 & 0.576 & 0.588 \\
PA-Score & \multicolumn{3}{c}{\multirow{3}{*}{--- \textit{Preserves L2} ---}} & 0.765 & 0.825 & 0.837 \\
SUE & \multicolumn{3}{c}{} & \textbf{0.161} & \textbf{0.215} & \textbf{0.225} \\
\textbf{KappaPlace-PT} & \multicolumn{3}{c}{} & 0.191 & 0.247 & 0.258 \\
\midrule
\rowcolor{gray!10} \multicolumn{7}{l}{\textit{Joint-Training Methods (Fine-tuned Backbone)}} \\
STUN & \textbf{90.0} & \textcolor{red}{\underline{95.2}} & \textcolor{red}{\underline{96.5}} & 0.423 & 0.475 & 0.487 \\
\textbf{KappaPlace-JT} & \textcolor{red}{89.3} & \textcolor{red}{\underline{95.2}} & \textcolor{red}{\underline{96.5}} & \underline{0.175} & \underline{0.230} & \underline{0.242} \\
\bottomrule
\end{tabular}}
\end{table*}

\begin{table*}[h]
\centering
\scriptsize
\caption{Detailed ablation study results for city-scale benchmarks: San Francisco XL and Pittsburgh 30k.}

\label{tab:ablations_city_scale}
\begin{tabular}{lllccccccccc}
\toprule
\textbf{Config. Name} & \textbf{Loss} & \textbf{Head} & \multicolumn{3}{c}{\textbf{SF-Test-v1}} & \multicolumn{3}{c}{\textbf{SF-Test-v2}} & \multicolumn{3}{c}{\textbf{Pitts30k}} \\
\cmidrule(lr){4-6} \cmidrule(lr){7-9} \cmidrule(lr){10-12}
 & & & \multicolumn{3}{c}{ECE@K $\downarrow$} & \multicolumn{3}{c}{ECE@K $\downarrow$} & \multicolumn{3}{c}{ECE@K $\downarrow$} \\
 & & & 1 & 5 & 10 & 1 & 5 & 10 & 1 & 5 & 10 \\
\midrule
GNLL-variant & GNLL & Agg. + Linear + Softplus & \textbf{0.141} & \textbf{0.082} & 0.090 & 0.230 & 0.330 & 0.357 & 0.333 & 0.409 & 0.427 \\
Linear-head & vMF & Linear + Softplus & 0.197 & 0.159 & 0.142 & 0.220 & 0.309 & 0.336 & 0.330 & 0.405 & 0.422 \\
KappaPlace-PT (Ours) & vMF & Agg. + Linear + Softplus & 0.208 & 0.106 & \textbf{0.078} & \textbf{0.160} & \textbf{0.263} & \textbf{0.290} & \textbf{0.218} & \textbf{0.294} & \textbf{0.312} \\
\bottomrule
\end{tabular}
\end{table*}

\begin{table*}[h]
\centering
\small
\caption{Ablation study results for environmental and long-term temporal benchmarks.}
\label{tab:ablations_environmental}
\begin{tabular}{lllcccccc}
\toprule
\textbf{Config. Name} & \textbf{Loss} & \textbf{Head} & \multicolumn{3}{c}{\textbf{AmsterTime}} & \multicolumn{3}{c}{\textbf{MSLS-val}} \\
\cmidrule(lr){4-6} \cmidrule(lr){7-9}
 & & & \multicolumn{3}{c}{ECE@K $\downarrow$} & \multicolumn{3}{c}{ECE@K $\downarrow$} \\
 & & & 1 & 5 & 10 & 1 & 5 & 10 \\
\midrule
GNLL-variant & GNLL & Agg. + Linear + Softplus & 0.326 & 0.157 & \textbf{0.101} & 0.243 & 0.307 & 0.346 \\
Linear-head & vMF & Linear + SoftPlus & \textbf{0.272} & 0.193 & 0.196 & 0.248 & 0.307 & 0.346 \\
KappaPlace-PT (Ours) & vMF & Agg. + Linear + Softplus & 0.309 & \textbf{0.148} & 0.105 & \textbf{0.155} & \textbf{0.200} & \textbf{0.238} \\
\bottomrule
\end{tabular}
\end{table*}

\begin{table*}[h]
\centering
\small
\caption{Calibration errors of query-level uncertainty scores: San Francisco XL and Pittsburgh 30k.}
\label{tab:unc_score_city}
\begin{tabular}{lllccccccccc}
\toprule
\textbf{Variant} & \textbf{Unc. Score} & & \multicolumn{3}{c}{\textbf{SF-Test-v1}} & \multicolumn{3}{c}{\textbf{SF-Test-v2}} & \multicolumn{3}{c}{\textbf{Pitts30k}} \\
\cmidrule(lr){4-6} \cmidrule(lr){7-9} \cmidrule(lr){10-12}
 &  & & \multicolumn{3}{c}{ECE@K $\downarrow$} & \multicolumn{3}{c}{ECE@K $\downarrow$} & \multicolumn{3}{c}{ECE@K $\downarrow$} \\
 &  & & 1 & 5 & 10 & 1 & 5 & 10 & 1 & 5 & 10 \\
\midrule
KappaPlace-PT  & $1/\kappa_Q$      & & \textbf{0.083} & \textbf{0.131} & \textbf{0.148} & 0.183 & 0.239 & 0.259 & \textbf{0.164} & \textbf{0.214} & \textbf{0.224} \\
KappaPlace-PT  & Eq.~\ref{eq:query_uncertainty} & & 0.093 & 0.163 & 0.182 & \underline{0.172} & \underline{0.231} & \underline{0.251} & 0.191 & 0.247 & 0.258 \\
\midrule
KappaPlace-JT & $1/\kappa_Q$      & & \underline{0.092} & \underline{0.148} & \underline{0.170} & \textbf{0.161} & \textbf{0.200} & \textbf{0.208} & \underline{0.172} & \underline{0.224} & \underline{0.233} \\
KappaPlace-JT & Eq.~\ref{eq:query_uncertainty} & & 0.097 & 0.154 & 0.174 & 0.220 & 0.262 & 0.270 & 0.175 & 0.230 & 0.242 \\
\bottomrule
\end{tabular}
\end{table*}

\begin{table*}[h]
\centering
\small
\caption{Calibration errors of query-level uncertainty scores: AmsterTime and MSLS-val.}
\label{tab:unc_score_environmental}
\begin{tabular}{lllcccccc}
\toprule
\textbf{Type} & \textbf{Uncertainty Score} & & \multicolumn{3}{c}{\textbf{AmsterTime}} & \multicolumn{3}{c}{\textbf{MSLS-val}} \\
\cmidrule(lr){4-6} \cmidrule(lr){7-9}
 & & & \multicolumn{3}{c}{ECE@K $\downarrow$} & \multicolumn{3}{c}{ECE@K $\downarrow$} \\
 & & & 1 & 5 & 10 & 1 & 5 & 10 \\
\midrule
Post  & $1/\kappa_Q$                  & & 0.343 & 0.186 & 0.136 & 0.111 & 0.150 & 0.171 \\
Post  & Eq.~\ref{eq:query_uncertainty} & & \textbf{0.310} & \textbf{0.145} & \textbf{0.108} & 0.074 & 0.116 & 0.138 \\
\midrule
Joint & $1/\kappa_Q$                  & & 0.366 & 0.184 & 0.152 & \underline{0.056} & \underline{0.094} & \underline{0.102} \\
Joint & Eq.~\ref{eq:query_uncertainty} & & \underline{0.339} & \underline{0.172} & \underline{0.133} & \textbf{0.040} & \textbf{0.071} & \textbf{0.076} \\
\bottomrule
\end{tabular}
\end{table*}

\subsection{Architectural and Training Details}\label{sec:appendix_impl_details}
\begin{figure}[h!]\centering\includegraphics[scale=0.65]{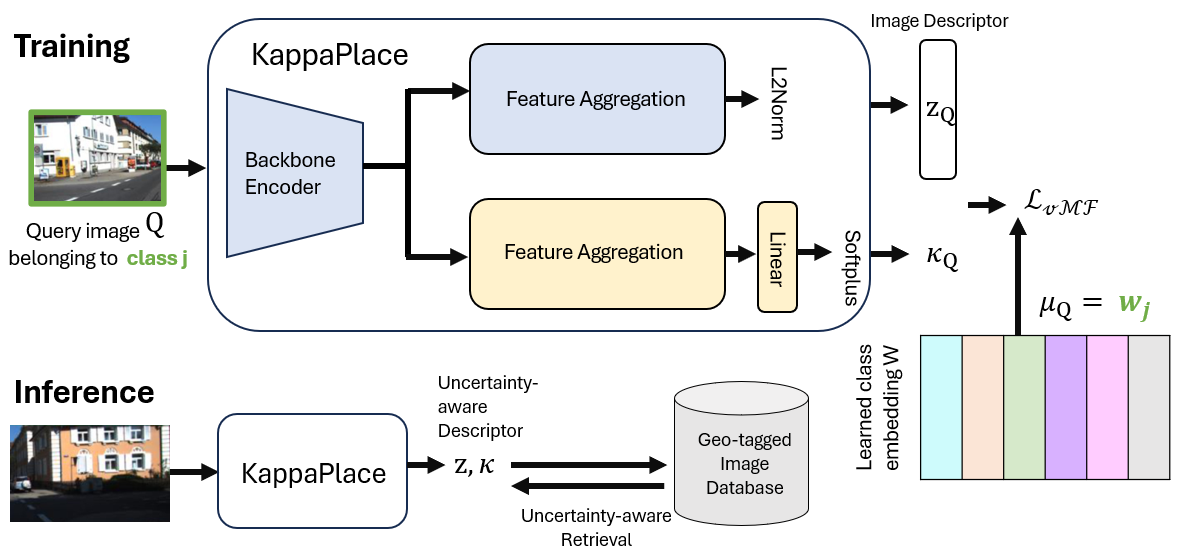}
\caption{Overview of the KappaPlace architecture during training and inference.}\label{fig:kappaplace_arch}
\end{figure}

The KappaPlace architecture, illustrated in Fig.~\ref{fig:kappaplace_arch}, extends a standard retrieval backbone with an auxiliary uncertainty branch that maps intermediate features to a concentration parameter $\kappa$. 
\paragraph{VPR Training.} For retrieval pre-training, we adopt the ResNet-50 setup from the official CosPlace repository, training for 50 epochs. We maintain standard hyperparameters: margin $M=10$, angle $\alpha=30^{\circ}$, $N=5$, and $L=2$. {For the Joint Training objective (Eq.~\ref{eq:total_joint_loss}), the balancing hyperparameter $\lambda$ was set to 0.01. Through empirical validation, we found that this value ensures stable convergence of both the discriminative and uncertainty-aware features without allowing the concentration estimation to dominate the gradient flow.} Optimization was performed using Adam and a batch size of 32.  Initially, the base model and class weights were trained with a learning rate of $10^{-5}$. For integrating the uncertainty variance head, two variants were evaluated: the first (KappaPlace-PT) froze the backbone and trained only the variance head with a learning rate of $10^{-3}$, while the second used joint training across the entire network with a learning rate of $10^{-5}$. 
All models produce a final descriptor dimension of 512. Primary results were obtained on an NVIDIA RTX PRO 6000 Ada (Blackwell Server Edition). For MSLS, images are resized to $512 \times 512$ pixels to account for inconsistent native dimensions.
\paragraph{Baseline Protocols.} For STUN, we follow the original protocol, training for 6 epochs. When normalized by iteration count, this corresponds to approximately 26 CosPlace epochs (43.8k iterations per STUN epoch versus 10k for CosPlace). We utilize a Sigmoid activation for the uncertainty output as per the original work. Notably, we removed the shift clamping used in the original STUN repository, as it was found to inhibit uncertainty learning in this architecture. 

For SUE, which exhibits an extreme long-tailed distribution, we utilize a log-transformed version (SUE-log) to ensure numerical stability during evaluation. We note that SUE's performance degraded without this transformation.

\paragraph{KappaPlace Training Strategies.}
\begin{itemize}
\item \textbf{Post-hoc (KappaPlace-PT):} We freeze the pre-trained backbone and batch normalization layers, training only the uncertainty head with an early-stopping objective based on ECE@1 (patience of 15 epochs).
\item \textbf{Joint Training (KappaPlace-JT):} We implement a phased early-stopping strategy. Phase 1 tracks Recall@1 until convergence (patience of 15 epochs); Phase 2 then continues optimization while tracking ECE@1 with refreshed patience.
\end{itemize}
\paragraph{Hardware and Reproducibility.} The primary experiments utilizing a ResNet-50 backbone were conducted on a high-performance server equipped with an NVIDIA RTX PRO 6000 Blackwell Server Edition GPU. Ablation studies and seed-stability analyses were performed using a ResNet-18 backbone. The ResNet-18 base was trained from scratch on an NVIDIA L4 GPU to ensure a consistent weight class metric, with early stopping triggered after 5 epochs of static Recall@1 performance.

\subsection{Inference and ECE@K Protocol}\label{subsec:appendix_ece}
To ensure robust ECE calculations, we implement a percentile-clamping protocol to manage outlier-induced tails. For vMF-based KappaPlace and the STUN head, we clamp both tails of the distribution at the 1st and 99th percentiles. For metrics with naturally one-sided distributions (e.g., L2-distance or SUE), we apply a one-sided clamp on the high-uncertainty tail only. Concentration values $\kappa < 1.0$ were floored to $1.0$ prior to ECE calculation. {While our training optimizes the loss, we implement an adaptive zoom-binning strategy during ECE evaluation to manage distribution shifts, consistent with the original STUN paper’s protocol.}
{To ensure reproducibility, we define the expected success level $\mathcal{C}$ for each bin $i$ in Eq.~\ref{eq:ECE_K}. Given $M$ actual bins (where $M = \text{num\_bins} - 1$), the expectation for the $i$-th bin (using 1-based indexing to align with our calibration plots, $i \in \{1, \dots, M\}$) is:
\begin{equation}
    \mathcal{C}(B_i) = \frac{M - i}{M - 1}
\end{equation} 
This mapping ensures that the most certain bin ($i=1$) is anchored to a perfect recall expectation of 1.0, while the most uncertain bin ($i=M$) is anchored to 0.0. Our final reported results utilize the percentile-clamping protocol during inference to ensure stability against extreme outliers.}

The aggregated results shown in the main text include all SF-XL tests sets (v1, v2, night and occlusion), Pitts30 and MSLS-Val. The Amstertime dataset was omitted since SUE cannot be properly evaluated on it, as explained in the main text.

Beyond global query reliability, we evaluate the calibration of individual query-reference pairs $(Q, R)$. For a total of $T = K \times N$ retrieved pairs, we bin each pair independently according to its match uncertainty $U_{Q \leftrightarrow R}$ (Eq.~\ref{eq:match_uncertainty}). In this context, the success rate of a bin, $\text{acc}(B_i)$, is defined as the fraction of query-reference pairs that are 
ground-truth positives. The match-level calibration error is calculated as:
\begin{equation}
ECE_{\text{match}@K} = \sum_{i=1}^{M} \frac{|B_i|}{T} \left| \text{acc}(B_i) - \mathcal{C}(B_i) \right|
\end{equation}

\subsection{Computational Efficiency Analysis}
To evaluate the practical applicability of the KappaPlace uncertainty module, we measure its computational footprint relative to the base CosPlace ResNet-50 architecture. We focus on three key metrics: inference latency, peak GPU memory consumption, and total model parameters. Inference latency is quantified by measuring the duration of a single forward pass for an image of size $512 \times 512$ pixels. To ensure statistical stability and mitigate the impact of transient hardware fluctuations, we average results over 200 consecutive runs following an initial 20-run warmup phase. Peak GPU memory is reported as the maximum memory allocated by the framework during a single-image inference pass. This methodology ensures a fair comparison of the additional overhead introduced by the secondary aggregation path and concentration prediction layer of the uncertainty module.  We report the additional inference time and memory required for KappaPlace in Table \ref{tab:efficiency}. Note that there is no difference between KappaPlace-PT and KappaPlace-JT at inference time.  

\begin{table}[h]
\centering
\caption{\textbf{Computational Overhead Analysis.} Inference latency, GPU memory, and model parameters introduced by the KappaPlace uncertainty module compared to the CosPlace ResNet-50 baseline.}
\label{tab:efficiency}
\small
\begin{tabular}{lcccc}
\toprule
\textbf{Metric} & \textbf{Baseline} & \textbf{KappaPlace}  \\
\midrule
Inference Latency (ms) & 5.10 & 5.44 \\
Model Parameters (M)   & 24.56 & 25.61 \\
Peak GPU Memory (MB)        & 148.90 & 161.0 \\
\bottomrule
\end{tabular}
\end{table}

\subsection{Extended Training Analysis}
The training durations for the proposed variants and baselines are summarized in Table~\ref{tab:training_times}. We train the CosPlace baseline for 50 epochs, following the original training recipe, which requires 26 hours. For STUN, we follow the original training protocol and train the student backbone for six epochs (requiring 47 hours)
after training the teacher baseline (total of 73 training hours).  KappaPlace-JT requires 73 hours to converge across both retrieval and calibration branches. KappaPlace-PT provides a modular alternative, requiring 23 hours to train the uncertainty head on top of the frozen backbone, resulting in a total cumulative training time of 49 hours. Importantly, while the PT variant is specifically designed to preserve the discriminative performance of the frozen backbone, the JT variant allows the model to learn uncertainty-aware features that occasionally outperform baseline model in specific benchmarks.

To verify that the performance gains of KappaPlace-JT are not merely a byproduct of its extended training duration (138 epochs), we conducted a controlled comparison by retraining the CosPlace (L2) baseline for additional epochs. Using an early stopping criterion based on Recall with a patience of 15, the baseline trained for a total of 114 epochs. As shown in Table~\ref{tab:l2_cosplace_app}, although this extended training yielded marginal improvements for the baseline, KappaPlace-JT consistently outperformed it across the majority of benchmarks, confirming that our prototype-anchored supervision provides architectural advantages beyond simple compute time. 
\begin{table*}[t]
\centering
\small
\caption{Training durations for different VPR configurations.}
\label{tab:training_times}
\begin{tabular}{l c}
\toprule
\textbf{Method} & \textbf{Training Time (Hours)} \\
\midrule
CosPlace (Baseline) & 26 \\
STUN & 73 (26 + 47) \\
KappaPlace-PT & 49 (26 + 23) \\
KappaPlace-JT & 73 \\
\bottomrule
\end{tabular}
\end{table*}
\begin{table*}[tbh!]
\centering
\small
\caption{Comparison of retrieval performance between KappaPlace-JT and the extended baseline (114 epochs).}
\label{tab:l2_cosplace_app} 
\resizebox{\textwidth}{!}{
\begin{tabular}{l ccc ccc ccc ccc}
\toprule
\multirow{2}{*}{\textbf{Method}} & \multicolumn{3}{c}{\textbf{SF-Test-v1}} & \multicolumn{3}{c}{\textbf{SF-Test-v2}} & \multicolumn{3}{c}{\textbf{Amstertime}} & \multicolumn{3}{c}{\textbf{MSLS-val}} \\
\cmidrule(lr){2-4} \cmidrule(lr){5-7} \cmidrule(lr){8-10} \cmidrule(lr){11-13}
 & R@1 & R@5 & R@10 & R@1 & R@5 & R@10 & R@1 & R@5 & R@10 & R@1 & R@5 & R@10 \\
\midrule
L2 (114 epochs) & 77.8 & 83.7 & 85.6 & 90.0 & 95.0 & 97.0 & 46.7 & \textbf{68.9} & 73.8 & \textbf{87.4} & \textbf{93.9} & \textbf{94.7} \\
KappaPlace-JT & \textbf{78.9} & \textbf{85.5} & \textbf{87.9} & \textbf{91.3} & \textbf{95.8} & 97.0 & \textbf{47.5} & 68.0 & \textbf{74.0} & 86.6 & 93.0 & 94.3 \\
\bottomrule
\end{tabular}}
\end{table*}

Furthermore, we examined the convergence of the STUN student model to justify our comparison protocol. As detailed in Table~\ref{tab:sfxl_final} and Table~\ref{tab:urban_longterm_final}, extending the training of the STUN student model from Epoch 6 (early) to Epoch 11 (late) does not result in significant gains in retrieval recall. In several instances, such as MSLS-val and SF-XL-v2, the calibration error ($ECE@K$) actually increases as training progresses, suggesting that the model begins to overfit or deviate from the teacher's manifold. This justifies the use of the Epoch 6 model as the most competitive and well-calibrated STUN baseline for our evaluations.
\begin{table*}[tbh!]
\centering
\small
\caption{Retrieval and Calibration results on SF-XL Benchmarks for the early (Epoch 6) and late (Epoch 11) training stages of the student model in STUN.}
\label{tab:sfxl_final}
\resizebox{\textwidth}{!}{
\begin{tabular}{l ccc ccc ccc ccc}
\toprule
\multirow{2}{*}{\textbf{Method}} & \multicolumn{6}{c}{\textbf{SF-XL-v1}} & \multicolumn{6}{c}{\textbf{SF-XL-v2}} \\
\cmidrule(lr){2-7} \cmidrule(lr){8-13}
 & \multicolumn{3}{c}{Recall@K $\uparrow$} & \multicolumn{3}{c}{ECE@K $\downarrow$} & \multicolumn{3}{c}{Recall@K $\uparrow$} & \multicolumn{3}{c}{ECE@K $\downarrow$} \\
 & 1 & 5 & 10 & 1 & 5 & 10 & 1 & 5 & 10 & 1 & 5 & 10 \\
\midrule
Epoch 6 & 75.6 & \textbf{84.1} & 86.1 & \textbf{0.288} & 0.355 & \textbf{0.371} & 86.8 & \textbf{93.5} & 95.3 & \textbf{0.350} & \textbf{0.417} & \textbf{0.432} \\
Epoch 11 & 75.6 & 83.6 & \textbf{86.2} & 0.289 & \textbf{0.352} & 0.372 & \textbf{87.0} & 93.3 & \textbf{96.0} & 0.383 & 0.447 & 0.470 \\
\bottomrule
\end{tabular}}
\end{table*}

\begin{table*}[tbh!]
\centering
\small
\caption{Retrieval and Calibration results on urban and long-term benchmarks for the early (Epoch 6) and late (Epoch 11) training stages of the student model in STUN.}
\label{tab:urban_longterm_final}
\resizebox{\textwidth}{!}{
\begin{tabular}{l ccc ccc ccc ccc}
\toprule
\multirow{2}{*}{\textbf{Method}} & \multicolumn{6}{c}{\textbf{amstertime}} & \multicolumn{6}{c}{\textbf{msls-val}} \\
\cmidrule(lr){2-7} \cmidrule(lr){8-13}
 & \multicolumn{3}{c}{Recall@K $\uparrow$} & \multicolumn{3}{c}{ECE@K $\downarrow$} & \multicolumn{3}{c}{Recall@K $\uparrow$} & \multicolumn{3}{c}{ECE@K $\downarrow$} \\
 & 1 & 5 & 10 & 1 & 5 & 10 & 1 & 5 & 10 & 1 & 5 & 10 \\
\midrule
Epoch 6 & \textbf{43.7} & \textbf{64.7} & \textbf{71.6} & 0.191 & 0.222 & \textbf{0.257} & 84.7 & \textbf{90.9} & 92.3 & \textbf{0.344} & \textbf{0.395} & \textbf{0.409} \\
Epoch 11 & 43.6 & 64.4 & 71.0 & \textbf{0.183} & 0.222 & 0.260 & 84.7 & 90.8 & \textbf{92.7} & 0.357 & 0.415 & 0.434 \\
\bottomrule
\end{tabular}}
\end{table*}

\subsection{KappaPlace Applications}
{Beyond the reported improvements, KappaPlace allows VPR to function as a reliable component of autonomous systems by providing dependable uncertainty scores that downstream pipelines can treat as probabilistic evidence rather than absolute truths. This is particularly valuable for autonomous navigation; for instance, in SLAM frameworks where environmental aliasing often leads to false loop closures, our match-level uncertainty provides a principled way to filter or prioritize constraints, preventing map corruption in ambiguous scenes. Furthermore, for robots operating in safety-critical domains, high-uncertainty signals can trigger conservative behaviors, such as reducing speed or cross-referencing secondary sensors, to enhance overall operational reliability.}


\end{document}